\documentclass[fleqn,10pt]{wlscirep}
\usepackage[utf8]{inputenc}
\usepackage[T1]{fontenc}
\title{DRL: Deep Reinforcement Learning for Intelligent Robot Control-- Concept, Literature, and Future}

\author{Aras Dargazany - arasdar@uri.edu}

\begin{abstract}
Combination of machine learning (for generating machine intelligence), 
computer vision (for better environment perception), and
robotic systems (for controlled environment interaction) 
motivates this work 
toward proposing a vision-based learning framework for intelligent robot control 
as the ultimate goal (vision-based learning robot). 

This work specifically introduces deep reinforcement learning as the the learning framework, 
a General-purpose framework for AI (AGI) meaning application-independent and platform-independent.

In terms of robot control, this framework is proposing specifically a high-level control architecture independent of the low-level control, 
meaning these two required level of control can be developed separately from each other.
In this aspect, the high-level control creates the required intelligence for the control of the platform 
using the recorded low-level controlling data from that same platform generated by a trainer. 

The recorded low-level controlling data is simply indicating the successful and failed experiences or sequences of experiments conducted by a trainer using the same robotic platform.
The sequences of the recorded data are composed of observation data (input sensor), generated reward (feedback value) and action data (output controller). 

For experimental platform and experiments, vision sensors are used for perception of the environment, 
different kinematic controllers create the required motion commands based on the platform application,
deep learning approaches generate the required intelligence, 
and finally reinforcement learning techniques incrementally improve the generated intelligence
until the mission is accomplished by the robot.

For deep reinforcement learning, deep Q networks (\textbf{DQN})~\cite{mnih2015human} using Convolutional Neural Network (\textbf{CNN})~\cite{lecun1998gradient} will be used for learning the Q network.
Deep Q Network approach is applied to the images acquired from our vision devices as input sensors
to select the maximum rewarded action (best possible move based on our experiences) for our output controller to generate the required kinematics motion.

\section*{keywords}
Advanced Cognitive Robotics, Control Architectures, Computer Vision, Deep Learning, Reinforcement Learning and Machine Learning.

\end{abstract}

\begin{document}

\flushbottom
\maketitle
\thispagestyle{empty}


\section*{Introduction}

This work specifically introduces deep reinforcement learning as the the learning framework, 
a General-purpose framework for AI (AGI) meaning application-independent and platform-independent.

In terms of robot control, this framework is proposing specifically a high-level control architecture independent of the low-level control, 
meaning these two required level of control can be developed separately from each other.
In this aspect the high-level control creates the required intelligence for the control of the platform 
using the recorded low-level controlling data from that same platform generated by a trainer. 

\begin{figure}[ht!] 
  \centering
  \includegraphics[width=\textwidth]{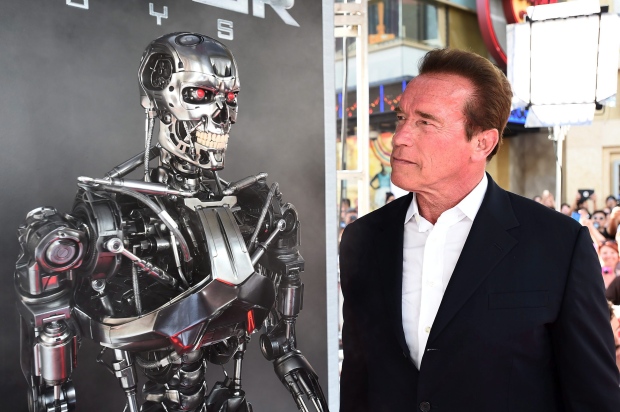} 
  \caption{Design and building a robot with biological inspiration has always been an ultimate goal in AI and robotics such as human to humanoid.}
  \label{figure:intro:me2amr}
\end{figure}

\begin{figure}[ht!] 
  \centering
  \includegraphics[width=\textwidth]{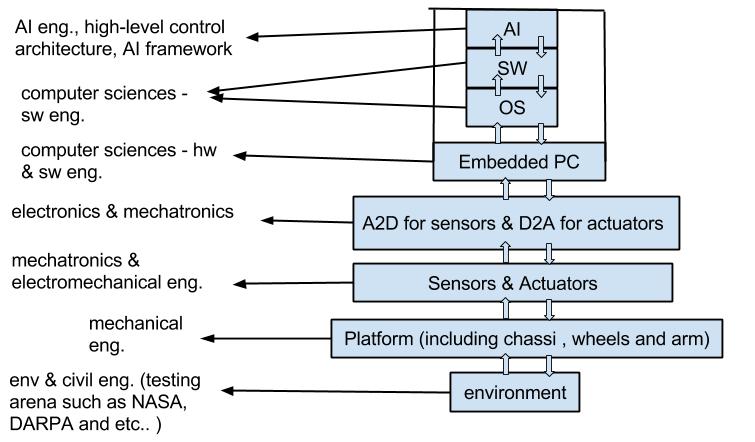}
  \caption{An AIR (Artificially Intelligent Robot), in above figure, shows how AI and Robotics are separated and benefiting from each other.
  It also shows how multidisciplinary is the background required for developing an AIR.}
  \label{figure:intro:air}
\end{figure}

\begin{figure}[ht!]
 \centering
 \includegraphics[width=\textwidth]{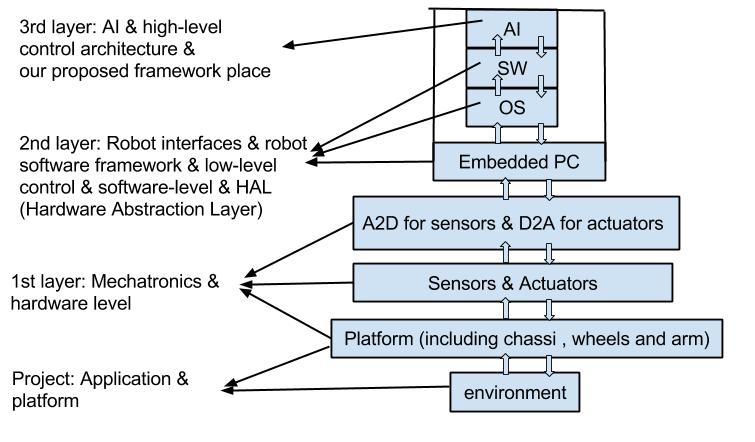}
 \caption{Three main layers required for developing an AIR.}
 \label{figure:intro:air_layers}
\end{figure}

The ultimate goal in AI and Robotics currently is to reach human-level control \ref{figure:intro:me2amr}.
Robotics and AI are among the most complicated engineering sciences and highly multidisciplinary, 
i.e. you should have good amount of basic knowledge in other sciences such as computer sciences, mathematics, electronics and mechatronics to get started with building an 
Artificially Intelligent Robot (AIR) as shown completely detailed in figures \ref{figure:intro:air} and \ref{figure:intro:air_layers}.

To introduce this work, we should start with the architecture of an AIR, 
more specifically how the Robot and AI should be combined together but at the same they should be separated (parallel development)
in order to know how they are benefiting from each other and how they can be developed independently from each other as well.
For better understanding and engineering of an AIR (design and build), we should have a better view of Robotics and AI as shown in figure \ref{figure:intro:air}; 

\begin{figure}[ht!]
 \centering
 \includegraphics[width=\textwidth]{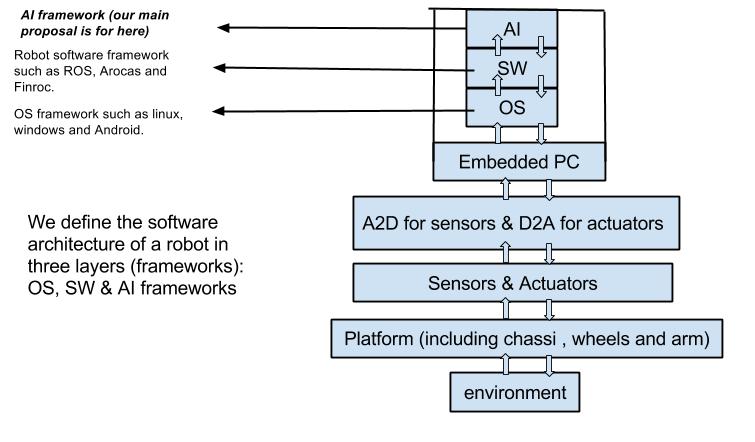}
 \caption{The proposed learning framework in this work is the general-purpose AI framework shown at highest level of software layers architecture.}
 \label{figure:intro:air_layers_sw}
\end{figure}

AIR is a combination of machine learning (for generating machine intelligence) and computer vision (for better environment perception) in terms of AI, and
robotic systems (for controlled environment interaction).
The AIR in this work is a vision-based learning framework for intelligent robot control as the ultimate goal 
as shown in figure \ref{figure:intro:air_layers_sw}.

The problem is simply the lack of a common general-purpose high-level control architecture independent of application and platform 
(i.e. the missing piece of puzzle is the AI framework shown in figure \ref{figure:intro:air_layers_sw}).
This work is aiming at approaching this problem by proposing a learning framework in a completely application- and platform-independent fashion (i.e. general-purpose).

In this work, we propose \textbf{DL (Deep Learning)} for generation of intelligence (mapping input sensor data to output controller data) 
along with \textbf{RL (Reinforcement Learning)} for improving the intelligence throughout the time (using the 4th dimension),i.e. learning from failed experiences or failures
to successfully accomplish the mission and achieve the ultimate goal.

\subsection*{Contributions of this work}
The outstanding contributions of this work will be as follows:
  \begin{itemize}
   \item Define Robotics and AI separately and emphasizing their bilateral relationship.
   \item Proposing a general-purpose AI framework as high-level robot control architecture, i.e. application- and platform-independent.
   \item Hopefully unifying the high-level robot control architectures and 
   uniting the efforts and researchers in this direction for a better understanding, more experiments and faster progress.
  \end{itemize}


\section*{Literature}
Looking at the literature, the robot control architectures design are rarely shown. 
This makes literature review hard, but
also, this can be an inclination toward a general-purpose framework as a missing puzzle.
In this chapter, we review the existing architectures for controlling machines since the early ages.

\subsection*{Robotics and Automation}
The robotics and automation simply started with the quest for controlling the machines automatically using sensors and controllers.
Automation had been center of attention for a long time since machines were manually controlled and used.
The existing literature is mostly as follows:
\begin{enumerate}
 \item Mechanical control such as gears and gear boxes
 \item Linear control from control theory using mechatronics and sensor feedback
 \item Non-linear control (this gave birth to intelligent control and shallow AI)
\end{enumerate}

\subsection*{Mechanical Control}
Mechanical solutions were applied to early machines for controlling them such as gears and gear boxes in the clocks and watches.

\subsection*{Control Theory, Linear Control and Mechatronics}
As the industry grows, the industrial machinery became more complex using electronics and ICs (Integrated Circuits), 
we were able to control machines with more flexibility and accuracy using conventional linear control systems using sensors for providing feedback from the system output. 

Linear control is motivated by Control Theory using mathematical solutions and specifically linear algebra implemented on hardwares using mechatronics, 
electronics, ICs (Integrated Circuits), and micro-controllers.

These systems were using sensors to feedback the error and were trying to minimize the error to stabilize the system output. 
These linear control systems were using mathematical solution known as linear algebra to drive the function that maps input to the output.
This field of interest was known as Automation and the goal was creating automatic systems.

\subsection*{Non-linear control}
Non-linear control became more crucial to drive the non-linear function (or kernel function) mathematically for more complicated task. 
The reason behind non-linearity was the fact that input and output had different and sometimes big dimensionality and the complexity could just not 
be modeled using linear control and linear algebra. This was the main motivation and fuel for the rise non-linear function learning or how to drive these functions.

\section*{Classical Robotics} 
By advancement in computer industry, non-linear control gave birth to intelligent control which is using AI for high-level control of the robot and systems. 
Classical robotics were the dominating approaches. 
These approaches were mostly application-dependent and highly platform-dependent.
Generally speaking these approaches were hand-crafted, hand-engineered and were addressed as shallow AI.

These architecture are also referred as \textbf{GNC (Guidance, Navigation and Control) architectures},
mostly composed of perception, planning and control modules.
Perception modules were mostly used for mapping the environment and localization of the robot inside the environment, 
Planning modules (also referred as navigation modules) to plan the path in terms of motion and mission,
Control modules for generating the controlling commands (controlling behaviors) required for the robot kinematics.

\section*{Probabilistic Robotics}
\label{sec:prob_rob}
The real game changer in robot control happened in DARPA 2005 that STANLEY was the only driver-less car for offroad challenge
using machine learning methods for navigation that could actually win the DARPA challenge in 2005 \cite{thrun2006stanley}. 
Sebastian Thrun in Probabilistic Robotics \cite{thrun2005probabilistic} introduces the new way of looking at robotics and how to incorporate \textbf{ML(Machine Learning)} 
algorithms for probabilistic decision making and robot control.

\textbf{DARPA 2007 - Urban challenge:}
Sebastian Thrun also introduced another driver-less car for urban challenge in DARPA 2007 as well which was also using ML techniques for probabilistic decision making and control. 
This car ended up finishing second among all the participants \cite{montemerlo2008junior}.

\textbf{Google self-driving car:}
After these two DARPA challenges, Sebastian Thrun started building self-driving car for Google using machine learning techniques 
which has proven to be working successfully in the two DARPA challenges mentioned before.

Also in terms of harmony in architecture theoretically and conceptually which in practice proved to be working very successfully, 
Sebastian Thrun's work really stand out in DARPA 2005 with Stanley platform \cite{thrun2006stanley}, 2007 with Junior platform \cite{montemerlo2008junior} 
and finally at Google with the self-driving car platform.

\begin{figure}[ht!]
  \centering
  \includegraphics[width=\textwidth]{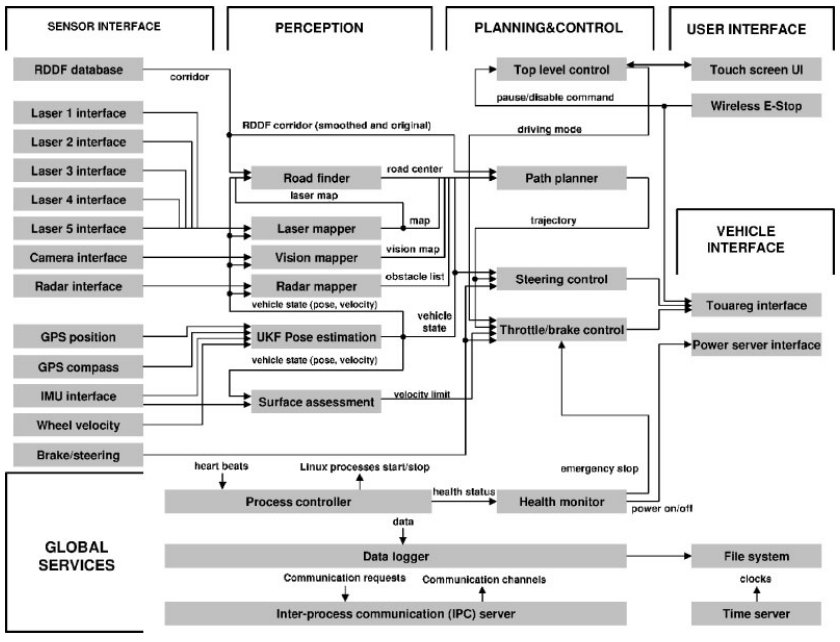}
  \caption{Stanley 2005 DARPA software architecture}
  \label{figure:sota:thrun:stanley05}
\end{figure}

\begin{figure}[ht!]
  \centering
  \includegraphics[width=\textwidth]{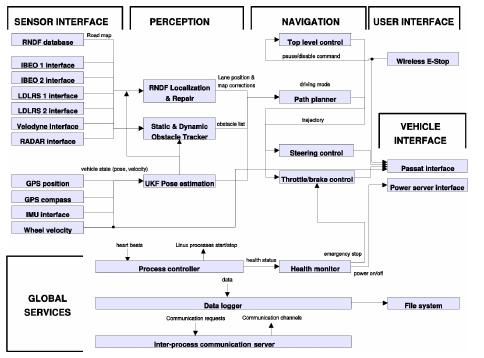}
  \caption{junior 2007 software architecture is very similar to Stanley's.}
  \label{figure:sota:thrun:junior08}
\end{figure}

\begin{figure}[ht!]
  \centering
  \includegraphics[width=\textwidth]{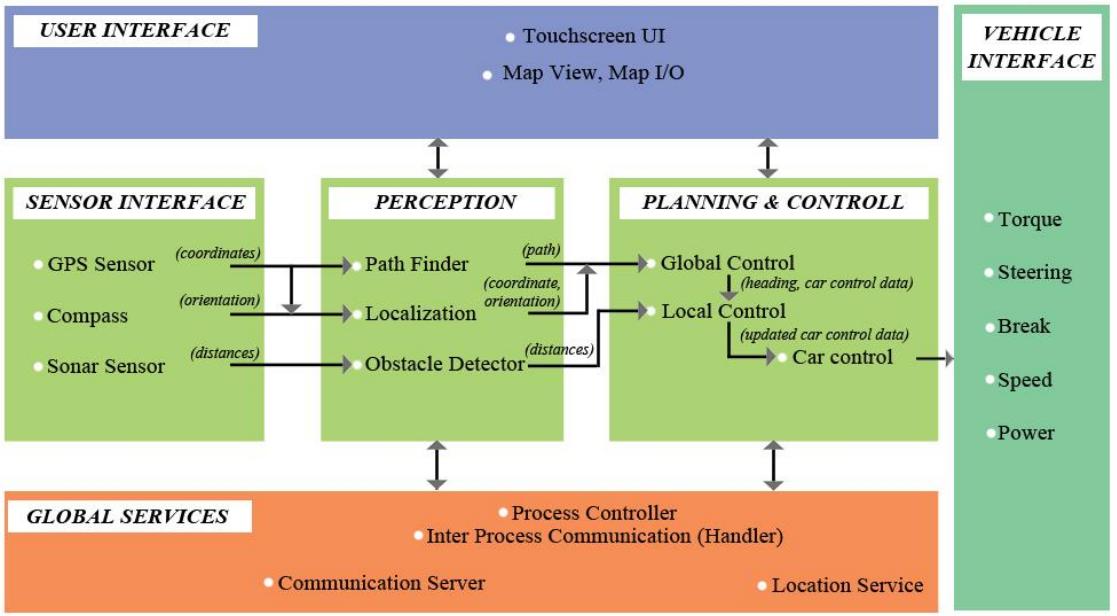}
  \caption{The Google self-driving car software architecture is highly similar to Stanley's and Junior's.}
  \label{figure:sota:thrun:google12}
\end{figure}

All these three very successful platforms are using the same conceptual control architecture design. 
By Looking at their software architectures (high-level control architectures) \ref{figure:sota:thrun:stanley05}, \ref{figure:sota:thrun:junior08}, and \ref{figure:sota:thrun:google12}, 
you can easily conclude that using the same architecture, we can build three different platforms which can successfully do the job.
This also highly motivated us to go toward proposing a general-purpose high-level control architecture.

These architectures are the best examples of classical robotics with the addition of the machine learning in planning/control (navigation/control) part. 
Sebastian Thrun impacted the field of robotics by adding machine learning from AI to the high-level control architecture (or system software architecture).

By looking at these three architectures, you can see how machine learning and computer vision have been used in a very successful way.
Aside from the interfaces, the main core architecture is composed of \textbf{Perception} and \textbf{Planning/Control or Navigation} 
(as you can see planning/control is equal to navigation in Sebastian Thrun's terminology).
The perception part has been fully regarded as a computer vision problem and has been solved using computer vision approaches, 
and on the other hand planning/control or navigation has been successfully solved using machine learning techniques and mostly \textbf{SVM}.

With the advances of machine learning algorithms in solving computer vision problems, machine learning as a whole (end-to-end) 
started to be taken a lot more seriously for intelligent robot control and navigation as an end-to-end approach for high-level control architecture (AI framework).
This gave a huge boost to cognitive robotics among the researchers and in the robotic community.

\section*{Cognitive Robotics and Deep Learning (DL): An Approach for Big Data Problem}
Applying machine learning as a framework for robot control architectures gave rise to the cognitive robotics.
Although there had been an enormous amount of attempts in this regard, they really did not succeed much
due to the lack of end-to-end learning framework before introduction of DL approaches.
They always had to incorporate some sorts of hand-crafted features or some pre- or post-processing conditions.

Since 2012, DL \cite{lecun2015deep} has brought excitement about AI general applications and 
perhaps a right way to approach intelligence in general \cite{jones2014learning}. 

Based on DL approaches, there have been a new trend in AI assistant, agents and systems from giant software companies, 
after observation of DL approaches performance on the big data problem due to their BIG data centers and servers. 
The current popular and well-known AI assistants are as following:
\begin{enumerate}
 \item OK-Google for Google inc. led by Geoffrey Hinton and Deep Mind inc., 
 \item Siri from Apple inc. led by VocalIQ co., 
 \item M from Facebook inc. led by Yann LeCun, 
 \item OpenAI inc. recently initiated and registered by Elon Musk, 
 \item Baidu inc. (Chinese version of Google) led by (Google brain project senior researcher) Andrew NG,
 \item Echo and Alexa from Amazon inc., and 
 \item Cortana from Microsoft inc. 
\end{enumerate}

\textbf{Deep Learning-based Robot and System Control:}
After introducing DL, state-of-the-art in AI was highly pushed in this direction 
since they over-performed other existing approaches in computer vision and speech recognition.
Also after Google self-driving car, twelve more companies in California were also registered as a start-up companies focusing only on producing self-driving cars, most of them using
the DL approaches. Below, we review the history of DL briefly:

\subsection*{Geoff Hinton: Introduction of Back-Propagation for Feed-Forward Neural Networks}
The main boost in reconsidering the use of Neural Networks was the introduction of Back Propagation algorithm as a fast 
optimization approach \cite{rumelhart1986learning}.
In one of his recent talks, he explained the biological foundation of back-propagation and how it might happen in our brain \cite{hinton2007backpropagation}.

At the same, he also proposed a parallel and distribute processing way of using these neural networks which actually makes it completely possible to
employ a big network for learning. This was almost the big bang of the idea of using Big Nets for Big Data problem in parallel distributed fashion \cite{rumelhart1988parallel}.

In almost the same year, he proposed a learning algorithm for Boltzemann machine (or networks) which are actually 
implemented on a vast distributed network of processors \cite{ackley1985learning} and \cite {hinton1986learning}.

In 2006, he proposed that Neural Networks are actually doing dimension reduction at all levels and that is the main power behind their layers \cite{hinton2006reducing}.
In the same year, he also proposed a learning approach for Distant Belief Networks in a big fashion and implemented on parallel distributed fashion \cite{hinton2006fast}.

The BIG BANG of DL happened on 2012 when he won the contest on ImageNet for object recognition \cite{krizhevsky2012imagenet} and the contest
in speech recognition \cite{hinton2012deep}. 
After this success, he was employed by Google and started using its BIG data centers for DL (BIG nets).

\subsection*{Bengio's work with LeCun: Introduction of Convolutional Neural Networks (CNN)}
Bengio's group with LeCun were the first to introduce the convolutional networks \cite{lecun1995convolutional_}, and consistently proposing them as best AI architecture~\cite{bengio2009learning}.
They successfully applied DL to OCR (Optical Character Recognition) for document analysis \cite{lecun1998gradient} 

\subsection*{Schmidhuber's work: Introduction of LSTM as Recurrent Neural Nets (RNN)}
LSTM architectures in Schmidhuber's lab was introduced for their short- and long-term memory \cite{hochreiter1997long}.
LSTM networks draw attention back to Recurrent Neural Networks.

He was also one of the figures, demonstrating the power of DL \cite{ciregan2012multi} specifically in image classification and categorization.
Specifically emphasizing on the simplicity but BIG (deep), he tried to demonstrate the power of the big-scaled simple networks in computer vision 
applications \cite{ciresan2010deep}.

An interesting review paper presents the complete historical trend of the DL approaches in Neural Networks \cite{schmidhuber2015deep}. 
This actually shows that these approaches have been existing for while but 
they were just not feasible due their enormous computational cost and they have been ignored.

\subsection*{Google brain project: Andrew NG contribution toward AGI and Large-Scale DL Experiment}
The biggest experiment even done on DL was performed by Google brain project 
\url{https://en.wikipedia.org/wiki/Google_Brain}, led by Andrew NG, one of the leading researcher in DL and Robotics~\cite{quigley2009ros}.
In 2012, they started implementing a BIG Dist Belief Network on a BIG cloud (parallel and distributed computing structure almost 16000 processors) using BIG data from YouTube data centers to crack the problem of AI and to prove the real power behind DL.
Results was published in this work~\cite{NIPS2012_4687}, recognizing the cat by watching BIG number of RAW videos, without any hand-engineering or any pre-  or post- processing of the videos.

As the result, Google employed Geoff Hinton 
and acquired his start-up DNN (short for deep neural networks) in 2013 as the main figure behind the introduction of DL.
Google also bought Deep Mind company in 2014 
in a heated race with Facebook. 
This company was known for the efforts toward AGI (General-purpose AI) using DRL (Deep Reinforcement Learning) approaches. 

Also due to the BIG processing power required for Google brain project, Google started also investing on the next generation of processors, known as 
Quantum computers that are highly capable of fast parallel processing.
This led to QuAIL project (\url{https://en.wikipedia.org/wiki/Quantum_Artificial_Intelligence_Lab}) joint with NASA.

Google acquired also Boston Dynamics as the most well-known robotics company 
in the world 
\url{www.bostondynamics.com}) 
for importing DL approaches into robots.

Having said all about Google brain, Deep Mind, QuAIL or Quantum AI Lab and Boston Dynamics, we can conclude that DL approaches, BIG computational power and robotics platform development
are all necessary for building a fast learning robot or true Artificially Intelligent Robot (AIR as also mentioned in introduction).

These changes motivated the researchers in robotics and deep learning community toward General-purpose AI framework using Deep Reinforcement Learning which will be explained 
more in details in the next section. 


\subsection*{Invitation for Machine Learning (AI) as AI framework}
One of the greatest work which is highly motivating our proposal is \cite{bagnell2015invitation} which is literally proposing learning by demonstration than
classical robotics. 
This proposal is highly motivated by this suggestion and this work \cite{bagnell2015invitation} is the main root and grounding of our work and 
will try to follow this path.

\subsection*{Advanced Robotics: Introduction of DRL (Deep Reinforcement Learning) approach}
DRL is a new born field of research which is the combination of deep learning and reinforcement learning as AGI (General-purpose AI) framework or artificial general intelligence framework.
Reviewing the literature in DRL is very important since it is growing really fast and it
is proved to be working in different application (application-independent) and on different platforms (platform-independent).
This is actually the main motivation for us although it is a very challenging and complicated topic.

\subsection*{Introduction of RL}
The works of Richard Sutton in introduction of RL \cite{sutton1998introduction} is indisputably the major motivation in the rush for using and employing these learning approaches.

\subsection*{Review on RL}
There is a very interesting review paper on RL approaches \cite{littman2015reinforcement_} which is published in Nature (\url{www.nature.com}).
The interesting point about this one is though that it really shows that deep learning and reinforcement learning papers are receiving an enormous attention nowadays.
People really started caring about how these approaches specifically deep learning approaches
are working and producing good results.
In \cite{littman2015reinforcement_}, reinforcement learning is introduced as a way to evaluate the feedback received from the environment and it will be used to improved 
the behavior.

It is specifically explaining the Markov Decision Processes in terms of specifying the settings and task such as state, action and reward chain for reinforcement learning.
The interesting part in this review is introducing self-aware learning in reinforcement learning which actually explains a lot about the power behind this approach. 
Also
at the end of this review, it is calling reinforcement learning a cognitive model which LITERALLY makes it somehow biologically inspired somehow or justify it biologically and 
some how explains it inside biological evolution framework.

Having read this work, really motivated to know more and more about reinforcement learning and how I can apply it to my work and general in robot control.

\subsection*{Review on RL in Robotics}
Reviewing the reinforcement learning approaches in robotics, proposing it as a framework in robotics as an application platform for RL experiments and case studies
are proposed in both \cite{kober2013reinforcement} and \cite{kober2012reinforcement} (these two works both have the same titles).

\subsection*{RNN with RL}
There is some interesting works by Eric Antonelo~\url{http://ericantonelo.com/} on robot navigation and localization or generally speaking complete robot control using learning approaches 
\cite{antonelo2015learning}.
He has been different machine learning techniques including reinforcement learning for different modules in robot control \cite{antonelo2011learning}. 
He is applying reinforcement using recurrent 
neural network which is usually called and addressed as reservoir inside his literature.

The great point about this work is approaching all the robot control problems such as perception and navigation (planning/control) using machine learning techniques. It is basically 
introducing machine learning as high-level control framework for intelligent robot control.

It also addresses localization problem with machine learning as well. In his PhD thesis \cite{antonelo2011reservoir}, he is completely showing the architecture of high-level robot control
and in this architecture, unsupervised learning, supervised learning and reinforcement learning all have used to address all robot navigation and control problem.
He also backs up his work by relating it to cognitive science, for biological inspiration behind the work.

I found this work really inspiring and uniquely interesting in a sense that it is proposing complete machine learning framework for intelligent robot control 
although it is not related to deep learning at all, specifically his work~\cite{antonelo2011reservoir} on robot navigation and localization using NN.
This work~\cite{antonelo2011reservoir} also has been conducted using machine learning-based robot navigation for all the modules using a simulated robot with a very limited sensors for localization and mapping the environment.

\subsection*{Advanced Robotics: Introduction of DRL by Google Deep Mind}
Deep Reinforcement Learning-based control or for short
DRL have been initially introduced and coined by a company called Google Deep Mind \url{www.deepmind.com}. 
This company started using this learning approach for the simulated agents in very old Atari games.
The idea is letting the agent learn on its own till it reaches the human-control level of gaming or maybe superior level.
Recent excitement in AI was brought by this DRL method \cite{mnih2015human} 
\textbf{DQN} in Atari games, simple simulated environment and robots 
for testing.

The latter proposed approach claims to outperform human-level control in the Atari games. 
This was a promising step toward AGI.
Afterwards, they moved to the 3D games and applied DQN to these games 
and it reached human-level control from scratch.

Their very latest achievement was building alphaGo \cite{silver2016mastering} and literally winning the contests with the giants of Go game,
starting from European champion and finally beating the world champion.

This  approach, \textbf{DQN}, has triggered some
researchers to look into the new theory of \textit{Learn-See-Act}
instead of the previous \textit{Sense-Plan-Act} \cite{scholkopf2015artificial} and \cite{littman2015reinforcement}.

Their main contribution until now is some how clear on their website. They have three main tabs for \textbf{Health}, \textbf{DQN} and \textbf{AlphaGo}.

\subsection*{Open-source work on RL and DL by OpenAI}
The introduction of OpenAI as well as its foundation and all that along
With the fast pace of DL approaches and by introduction of DRL approaches for AGI \url{https://en.wikipedia.org/wiki/Artificial_general_intelligence},
Some of the wealthy independent thinkers such as Ellen Musk \url{https://en.wikipedia.org/wiki/Elon_Musk} started procrastinating the actual future applications of AGI
and started somehow worrying it.

This gave birth to a foundation with the aim of open-sourcing the AI technology and
get people involved in the actual development, gathering all the important minds in the field and conducting open-source research. 
OpenAI \url{www.openai.com} was born, mainly focusing on RL and DL approaches.

This This work~\cite{HeinrichS16} proposing DRL as self playing approach in non-perfect environment in terms of the provided information, specifically in games as the environment and for agents.
Again DRl as self-playing, self-learning and self exploring approach experimented in games as the most primitive simulated environment and agents.

In terms of DRL, this work~\cite{wangdeep} is demonstrating "Case Study with Standard RL Testing Domains" which provides a great depth and intuition about
DRL as a general-purpose framework. 

This work~\cite{BaramZM16} explains more about how DRL is exploring different models (internal model) and somehow trying to understand the power behind this approach by analyzing it more 
in different parts and how it comes up with new model.

This work~\cite{FoersterAFW16a} is a good example of applying DRL for multi-robot communication. Centralized learning and decentralized execution of commands on each agent platform.
This shows the potential of DRL to be applied to a team of robot and learning how to communicate and share the information for better decision making.

This work is another work with Prof. Abbeel participation:
Another work for continuous control outputs and decision making is \cite{DuanCHSA16}. It shows DRL framework for generating continuous control command or action as a  regression method
for function fitting (classification used for function approximation with discreet output action).

This work~\cite{Peng:2016:TLS:2897824.2925881} shows the application of DRL on terrain traversability based on kinematics of the agent for locomotion. 
This is very well related to my previous work \cite{Dargazany14} trying to solve the traversability problem using a hand-engineered features such as surface and points normals.

This work~\cite{ZhaoE16} is an application of DRL as an end-to-end framework for NLP (Natural Language Processing) and also understanding along with speech recognition.
The DRQN is proposed as the DRL framework specifics which is the combination of DRNN (Deep RNN) and Q-network.

This work~\cite{DurugkarRDM16} proposed DRL as a powerful framework for learning policies to map BIG sensor input data to the output actions in different domains such as Atari games.
This work is concentrating on macro-actions as the output controlling commands in 2600 different Atari games.
They mainly try to address the shortcomings of DQN \cite{mnih2015human} and reason about the problem of sparse rewarding signals.

This work~\cite{Merrill16} is a great example of policy learning so-called P-learning in the paper in DRL framework. 
This is simply explaining how a DL approach can be inside RL setting using negative examples.

In terms of transfer learning, this work \cite{RusuRDSKKPH16} claim the first successful experimentation continual learning.
This work in terms of transfer learning or continual learning is very important.
Briefly speaking, they claim the addition of memory to the DRL framework successfully.

This work~\cite{KempkaWRTJ16} is applying DRL framework to simulation framework of the environment and the agent based on a 3D game called Doom and reporting its results.
This simulation framework is based one-person view which makes much more realistic than Atari games.
It also shows the architecture of the CNN used for DL along with Q-learning as RL approach in this aspect.
This work~\cite{TesslerGZMM16} is proposing a lifelong DRL framework including transfer learning as well for reusing the learned skill for the next task.

\section*{Industrial Robotics: DRL (Deep Reinforcement Learning) applications and requirements}

This chapter is specifically introducing the important requirements and challenges of DRL framework application to industrial robotics,
review the latest literature related to the works done in this direction, and 
propose the important remaining frontiers.

\subsection*{Some comments from engineering perspective vs computer science perspective}
\label{sec:intro}
As I am in engineering and not in computer science, the engineering
application of DRL concept, specifically DQN application on Atari game as proposed in DQN paper~\cite{mnih2015human} should not just be a \textit{"toy example" like learning
Pong!}, but a realistic scenario that resembles practical needs, i.e. in
automated production (engineering applications mainly).
This also improves chances that your concept might
later be used in industry, and improves the chances for a later position or application of DRL in
industry. 
And I understand that engineers are interested in such realistic
scenarios (otherwise, a computer science institute which would be the better
option for DRL development at this point).

Therefore, Ph.D. theses or researcher works must contain or add a chapter or sub-chapter on the
requirements for robots in industry.
Important requirements are, among others as following:

\subsection*{Accuracy}
\textbf{- accuracy: }
Robot movements have to be within certain accuracy requirements.
For grinding, e.g., this is in the magnitude of micrometers. 
For painting,
this is in the magnitude of a couple of millimeters.

\textbf{Some comments addressing the issue:}
-> accuracy can be achieved based on examples, i.e. the higher accuracy, the more examples;
-> we do not start from scratch but we start from an expert experience (initializing by learning-from-expert=demonstration) manually controlling robot and provide us with good samples of training;
-> expert feedback should be provided to robot as an external rewarding system, i.e. an expert should be constantly watching the robot and grade the robot performance such as deep reward learning-by-demonstration; 

\subsection*{Safety}
\textbf{- safety:} A collision of a robot arm (if DRL applied to grasping and robotic arm manipulation) with anything in the robot cell results
in damage and (more severe) loss of production during production time. 
This must be avoided by
suitable mechanisms, i.e. safety mechanism such as CQL (conservative Q-learning), safety critic and adversarial RL approaches for robustness. 
\textbf{-> The DRL mechanism} is learning by reinforcement signal (reward or punishment) since we are saying this approach is general-purpose AI or AGI. 
So we can not use DRL for one task and use some other approach for another task. 
This is against our claim, i.e transferability and task-agnostic issue.

\textbf{some comments addressing the issue:}
-> this should also be approached using training, examples and feedback;
-> another important parameter here can be extra sensors but using more sensors also bring us the issue of multi-modal learning and fusion inside the network;

\subsection*{Robustness or robust RL or adversarial RL or deep adversarial RL}
\textbf{- robustness:} The robot will have (and usually are installed or used) to operate in an environment with changing conditions. 
Even though industrial production has provided clearly structured environments or controlled environment (much clearer than road traffic or outside world or wilderness), some conditions might change, e.g. the lighting conditions. 
The intelligence robotic system must be able to cope with this. 
This work~\cite{zhang2015towards} shows that their DRL approach works
fine as long the input data stems/comes from simulation (related to domain transferability from sim-to-real). 
Similarly, the often-reported applications to Atari games~\cite{mnih2015human} take ideal pixel information as a basis.

\textbf{Some comments addressing this issue:}
-> lighting conditions affect an image and that is why the camera of the image can provide enough information as far as the lighting condition is letting the robot to see something. 
we can NOT expect in a very bad lighting condition that robot can not see any thing or the images are not clear at all, that the robot is able to operate;
-> this DRL approach do not need the pre-engineered environment and nor pre-engineered data since pre-engineered environment provide pre-engineered data.
As far as the data is understandable, distinguishable and informative for the robot, the DRL learning approach should be able to adapt itself with the changes.
-> For robustness, extra sensors are also required, e.g. laser scanners or infra-red cameras for geometry features acquisition for the sake of lighting conditions problem which highly affects the robot camera images.
Using extra sensors, highly affects the accuracy, safety and robustness. 
That is why fusion or multi-modal learning is a very important problem which is actually not being very well investigated and researched in DRL or AGI community.

Regarding these three problems, Deep Reinforcement Learning (DRL) approach is supposed to be general-purpose, i.e. application (environment), platform (machine) and task(the actual job) independent, also called task-agnostic.

\subsection*{Flexibility}
\textbf{- flexibility:} Robots are already frequently used for repetitive tasks, such as car body welding. 
However, in today's economy, lot sizes shrink, and products become more and more individual (think of sport shoes, for example), and therefore robots must be flexible to be adapted to new product variants. 
For large lot sizes, it was feasible to teach the robot a certain
movement (either by moving it with a joystick or by moving it manually with
a force sensor at the tool center point). 
This becomes inefficient and costly for small lot sizes. 

\textbf{Some comments addressing the issues:}
-> we have to provide robot with samples of our work. 
we should be able to tell robot what we want him to do.
Assuming that we can do such thing ourselves. 
If we can not do something, we can not expect the robot to do this on its own with few examples.
If we want the robot to do such thing, it needs to start from scratch, with a complete try and error and our feedback. 
This is very costly.
We can also start with a point that somehow we show robot an approximate sample of what we want or the task, and then let the robot robot start practicing it
with our feedback until its performance is good enough for us.
This is usually mentioned in the literature as learning-from-expert demonstration, imitation learning, behavior cloning, and also deep reward-learning-by-demonstration.
GAIL is an important work in this respect which is combining GAN 
for imitation learning of the training data distribution. 

\textbf{Possibly important questions in this regard:}
\begin{enumerate}
 \item How do I think we can address such requirements?
-> they are individually addressed above.

 \item Is the need for flexibility a hindrance/obstacle for reinforcement learning (DRL) approach? 
-> This is not a learning approach. 
DRL is LITERALLY the ONLY serious candidate currently to reach true AGI (General-purpose AI or artificial general intelligence) based on the literature and the results~\cite{mnih2015human, silver2016mastering}.
This approach is the only approach which could have possibly satisfy the flexibility to a great-extent.

 \item Or is this learning approach (DRL) a suitable solution to the request for flexibility?
It is the ONLY possible approach available. 
No other choice.
If we are in science, we should respect the scientific works and results~\cite{mnih2015human, silver2016mastering}. 
That is why best on the published literature,
the best state-of-the-art result has been reported with this approach~\cite{mnih2015human, silver2016mastering}.
\end{enumerate}

These questions need to be answered before we can start to apply the concept of DRL
to a robotic use case. 
Otherwise, it might be a waste of time according to industry since DRL has not been officially adopted by industry. 
-> As I have been looking for a research/PhD position supporting this research topic, I found out one thing for sure:
There are not a lot of groups and labs which are investing in this topic.
They are either not interested or do not believe in this approach.
As you probably might notice, this is a very new topic as I also discussed it in literature section (previous section) 
Deep learning has been taken seriously since mid of 2012 but Deep Reinforcement Learning has been just around since 2015 since the publication of DQN paper in Nature~\cite{mnih2015human} and also AlphaGo paper was its first practical application~\cite{silver2016mastering}.
By the way, DQN~\cite{mnih2015human} work was purely done very primitive Atari games by a serious AI company, Deep Mind (not a computer science group). 
Atari games used to be regarded as toy examples of the simulated environment and simulated agent but they were perfectly used in this scenario to prove a concept of a 
general-purpose approach for a lot of different games.

-> As you probably already noticed, I have not done many experiments on this platform of DRL approaches due to lack of funding available in this area specially in academia, and I am completely new in implementation of these approaches and applying them, 
but I have thoroughly reviewed the literature as I am writing this paper and I have a very detailed understanding of these approaches and their power and their current problems. 

\section*{Literature related to the industrial applications of DRL}

In terms of experiments and application of DRL would almost look like this work~\cite{zhang2015towards} which is applying DQN~\cite{mnih2015human} for intelligent robot control in discrete action space.
This work is important ONLY because it is showing the first application of DQN~\cite{mnih2015human} to a practical example of robotics. 

Undoubtedly, the most inspiring works in this aspect for considering DRL framework for intelligent robot control are being done in Pieter Abbeel's group \url{https://people.eecs.berkeley.edu/~pabbeel/} for applying AI, specifically deep learning and more specifically deep RL or DRL to the robots and making them learn.
The very important point on his page \url{https://people.eecs.berkeley.edu/~pabbeel/} is offering two new courses:
\begin{enumerate}
 \item Advanced Robotics \url{https://people.eecs.berkeley.edu/~pabbeel/cs287-fa15/}
 \item DRL \url{http://rll.berkeley.edu/deeprlcourse/}
\end{enumerate}
If check out this link \url{https://www.youtube.com/user/pabbeel/feed}, you will see the latest demonstration of DRL applied to robotics.
Also you can listen to his talk~\url{https://www.youtube.com/watch?v=evq4p1zhS7Q} which I found the latest and the most important one in terms of addressing the issues and my comments mentioned above along with his new podcast named Robot Brains. 

Another industrial application of DRL is presented in this work~\cite{Guenther20161}. 
DRL introduces a cognitive control architecture for robotics (or intelligent robot control or cognitive robot control) used for laser welding as a complex industrial process.
This is a perfect example of applying DRL to an industrial application with high importance in terms of the final accuracy, robustness, safety and flexibility.
This work~\cite{GUNTHER2014474} shows the earlier work in this respect done around 2014 and almost at the beginning of DRL big boost~\cite{mnih2015human}. 
This is the earlier work~\cite{GUNTHER2014474} and as an incremental work~\cite{Guenther20161}, it is explaining the continuation of this work~\cite{GUNTHER2014474} in this aspect.

There is a very similar work~\cite{Zhang201553} with the same application and also using ANN with BP (Back Propagation)~\cite{rumelhart1986learning} for learning and optimized using GA (Genetic Algorithm).
This work is basically proposing a DL (Deep Learning) approach using PCA for feature extraction layer as an unsupervised learner and ANN using BP optimized with GA for the classification layer (function approximation) as supervised learning.

This work~\cite{Mocanu2016646} is also an interesting application of DRL framework to energy domain in smart grids for complex decision making processes. 
Although this work is applied to a dataset, but still the application domain is very new and problematic.
Two RL approaches have been used with deep DBN (Deep Belief Network)~\cite{hinton2006fast} as DL approach.
The very interesting point of this work is the addition transfer learning on top the DRL to transfer the learned knowledge to other tasks.

This work~\cite{Deng16_7407387} is a perfect example of DRL applied to other domains such as finances. 
RNN and RL are used for DRL also they introduced RL and RNN as both being biologically inspired approaches. 
Also the interesting aspect is that they introduce DRL as a framework, composed of DL for temporal difference learning of the market, and RL makes it possible to explore the unknown environment to collect as much reward as possible (maximized the total expected reward). 
The proposed DRL framework in this work~\cite{Deng16_7407387} is applied and verified in both stock and commodity market.

Another similar application in the residential load control is~\cite{ClaessensVR16}.
This work is proposing DRL framework for high-dimensionality problem (Big Data problem) with partial observability (occlusion or lack of data problem), also known as partially observable Markov decision processes environment (POMDP).
CNN for learning the Q-network as DQN as DRL framework is proposed.

This work~\cite{Cuayahuitl16} is demonstrating a very simple DRL framework for speech to action. 
It is explained and written in a very simple language for understanding the complication of DRL.
This is also another general application domain of DRL such as restaurant domain applications for creating an intelligent dialogue system for interacting with customers.

Yamins and DiCarlo's work~\cite{yamins2016using} is published in Nature journal and it is completely discussing the strong biological foundation of DL as it is name as HCNN (Hierarchical CNN).
It is beautifully comparing computational and cognitive neuroscience in terms of their conclusion of visual cortex and other part of brains working the same as the recent DL approaches as computational neural networks.
It introduces DRL as goal-driven deep networks and back it up biologically throughout the whole paper very strongly.

This work~\cite{2016arXiv160802239J} is a great work dealing with the precision of DRL framework applied to the robot manipulation task.
Prof. Levine is an active researcher in this field in terms of application of DRL frameworks to Robotics and other possible domains as shown below:
\begin{enumerate}
 \item \url{http://iplab.dmi.unict.it/icvss2016/Abstracts/Levine.pdf}:
 Deep Learning for Robotics-- Learning Actionable Representations;
 \item \url{https://archive.org/details/Redwood_Center_2015_11_04_Sergey_Levine_and_Chelsea_Finn} 
 : This is an important lecture presenting his work.;
 \item \url{https://sites.google.com/site/brainrobotdata/home}
 : this is the link the \textit{Google Brain Robotic Grasping project} in which he is the leading researcher;
 This project is very important in a sense that it shows how much Google is investing in such DRL frameworks at the moment specifically for robotic domain applications.
 \item \url{https://research.googleblog.com/2016/03/deep-learning-for-robots-learning-from.html}: this shows the demonstration of robotic grasping and how they are practicing.
 It is literally like an elementary \textbf{school} of robots learning how to grasp and manipulate the objects. 
 \item The latest publication related to the grasping school for robots is shown in these two works \cite{LevinePKQ16} and \cite{FinnGL16}.
 This literally discusses the large-scale application of DRL framework for schooling the robots how to do an accurate, complex, robust grasping and manipulation of the objects using ONLY one RGB-D camera.
 \item This work of his~\cite{levine2016end} is a very important work in terms of end-to-end training for a very deep network for control policy and its application to many manipulation tasks.
 The DRL framework in this work is an end-to-end network using vision and generating the torques command for different complex task in manipulation domain.
\end{enumerate}

Based on the above links, we can simply conclude that Google and Nvidia are heavily investing into the idea of DRL framework, specifically in robotics domain and graphical domain for intelligence robot control using real and simulated environments.

\section*{Proposed potential important future frontiers} 
This section proposes some important future frontier to reach machine intelligence (General-purpose AI or AGI) similar to human intelligence (biological intelligence).

\textbf{Multi-modal fusion:}
Multimodal deep learning is a solution to this.
Ton of sensors with high dimensional data acquisition such as Velodyne, laser scanners, cameras cause the BIG DATA problem which can be solved by BIG neural network (BIG equals to deep neural network).
Also different controllers on the other end with multiple arms and complicated wheel kinematics.
This is another frontier: can we control them all using the same algorithm.
This is another end (output end) of the fusion problem.
Prof Abbeel PhD thesis - apprenticeship learning for robot control using inverse RL was one of these works. 

\textbf{Reinforcement learning} includes emotion, - satisfaction, - reward,
and goal, accuracy and robustness all at once.
Although there is a vast literature behind it BUT still the best working DRL approach til now is AlphaGo~\cite{silver2016mastering} which is in fact composed of two neural networks: 
one is an action evaluator or critic and another one is an action selector or actor (actor-critic architecture).

\textbf{In order to reach true autonomous level self-reward} generation is essential or also self-supervision or self-supervised learning (SSL) as Yann LeCun phrased it.
Self-reward generation highly depends on the memory of experiences including both successes and failures.
This is how I divide autonomy from intelligence and from automation as following:
autonomy: self-reward generation, 
intelligence: self-decision making,
automation: self action generation;

\textbf{one important suggested future direction:}
-invitation for imitation learning or mimicking behaviors (behavior cloning) learning-by-demonstration as proposed and motivated by this work~\cite{bagnell2015invitation}:
This is/was a PhD-level project idea to use vision for teaching by demonstration.
For example, around combining deep neural networks and reinforcement learning, DRL.

\textbf{Another research problem that might be interesting is the following:}  
In the future, we can expect to see robots deployed in houses and age-care facilities. 
Although these robots might come with a large set of built-in skills, there will be tasks for which the user will want to instruct the robot how to execute a new task without having to write new code.

\textbf{I also should add one other more very important point:}
DRL is LITERALLY the ONLY serious candidate currently to reach true AGI (General-purpose AI).
What I mean by general-purpose is application-(environment), platform-(machine) and task-(the actual job) independent.

\textbf{One possible approach to this problem is to teach by demonstration.}
However, there is a need for innovative methods that will work with very few examples (few shots learning or learning by few examples).
This is a fundamental problem with a large number of applications absolutely because it is application-independent approach (general-purpose approach).

\textbf{An example robotics platform specifically for DRL research frontiers:}
This one is about vision based control of a fast KUKA arm (KR6 R 900 Sixx Agilus) capable to learn how to play table tennis as a robotic platform for DRL research.
Everything is already installed on it (e.g. OS, libraries, frameworks).
This KUKA robotic platform has an arm with table tennis bat mount, a ball tracking system with 4 high speed cameras (out of which currently only 2 are used but this could be extended upon request and requirement), a ball spin detection system with a fifth high speed camera, a robot safety system with 2 SICK laser scanners (LIDAR), special high brightness LED ceiling lighting, a table tennis ball throwing machine/robot, etc.
While the first steps will be to control the robot to reach a desired ball intersection point with desired bat angle and velocity, the further steps shall involve learning algorithms, either deep reinforcement learning or (preferred) deep neural network learning,e.g. with variants of LSTM networks~\cite{hochreiter1997long}, for intelligent robot control.


\end{document}